\documentclass{bmvc2k}

\usepackage{booktabs}
\usepackage{multirow}

\usepackage{amssymb}
\usepackage{pifont}
\usepackage{algorithm}
\usepackage{algpseudocode}
\usepackage{multicol}

\newcommand{\methodname}[1]{\mbox{\emph{#1}}}

\title{Online Adaptation of Convolutional Neural Networks for Video Object Segmentation}

\addauthor{Paul Voigtlaender}{voigtlaender@vision.rwth-aachen.de}{1}
\addauthor{Bastian Leibe}{leibe@vision.rwth-aachen.de}{1}

\addinstitution{
 Computer Vision Group\\
 Visual Computing Institute\\
 RWTH Aachen University\\
 Germany
}

\runninghead{Voigtlaender, Leibe}{Online Adaptation for Video Object Segmentation}

\def\eg{\emph{e.g}\bmvaOneDot}

\def\etal{\emph{et al}\bmvaOneDot}

\def\ie{\emph{i.e}\bmvaOneDot}
\def\cf{\emph{cf}\bmvaOneDot}

\newcommand{\PAR}[1]{\vskip4pt \noindent {\bf #1~}}
\newcommand{\PARbegin}[1]{\noindent {\bf #1~}}

\begin{document}

\maketitle

\begin{abstract}
We tackle the task of semi-supervised video object segmentation, \ie segmenting the pixels belonging to an object in a video using the ground truth pixel mask for the first frame. We build on the recently introduced one-shot video object segmentation (\methodname{OSVOS}) approach  which uses a pretrained network and fine-tunes it on the first frame. While achieving impressive performance, at test time \methodname{OSVOS} uses the fine-tuned network in unchanged form and is not able to adapt to large changes in object appearance. To overcome this limitation, we propose Online Adaptive Video Object Segmentation (\methodname{OnAVOS}) which updates the network online using training examples selected based on the confidence of the network and the spatial configuration.
Additionally, we add a pretraining step based on objectness, which is learned on PASCAL. Our experiments show that both extensions are highly effective and improve the state of the art on DAVIS to an intersection-over-union score of $85.7\%$.
\end{abstract}

\section{Introduction}
\label{sec:intro}

\begin{figure}[t]
\begin{center}

\newcommand{\myscale}{0.08}

\begin{minipage}{0.12\textwidth}
\begin{center}
\vfill
\small
un-adapted\\
baseline
\vfill
\end{center}
\end{minipage}
\begin{minipage}{0.87\textwidth}
\includegraphics[scale=\myscale]{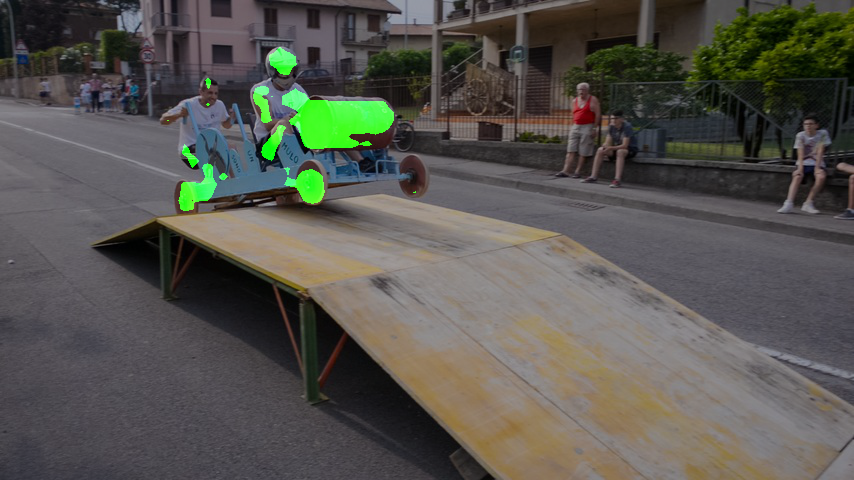}
\includegraphics[scale=\myscale]{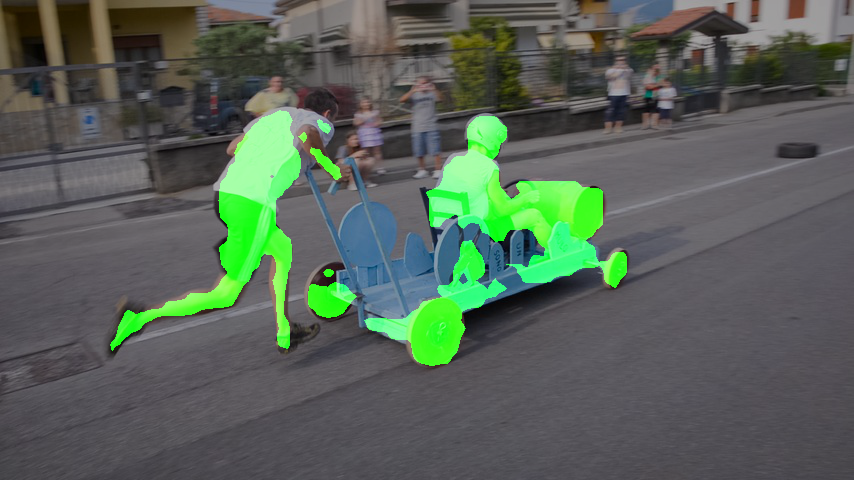}
\hspace{0.1cm}
\includegraphics[scale=\myscale]{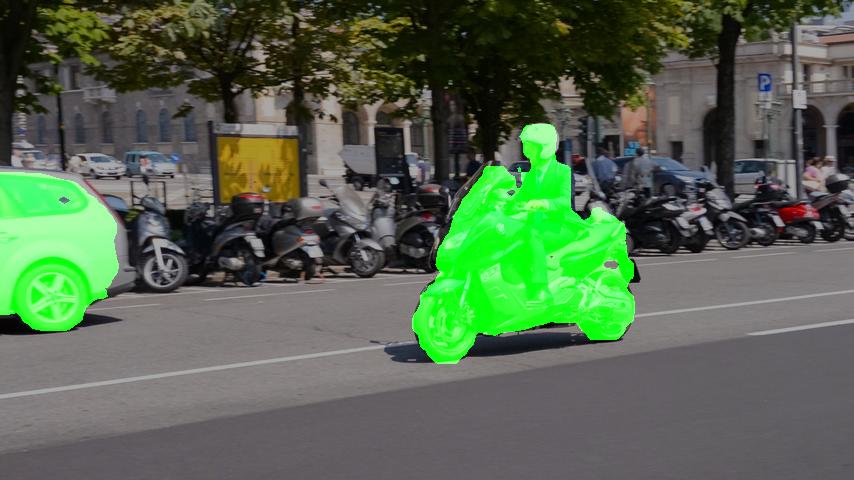}
\includegraphics[scale=\myscale]{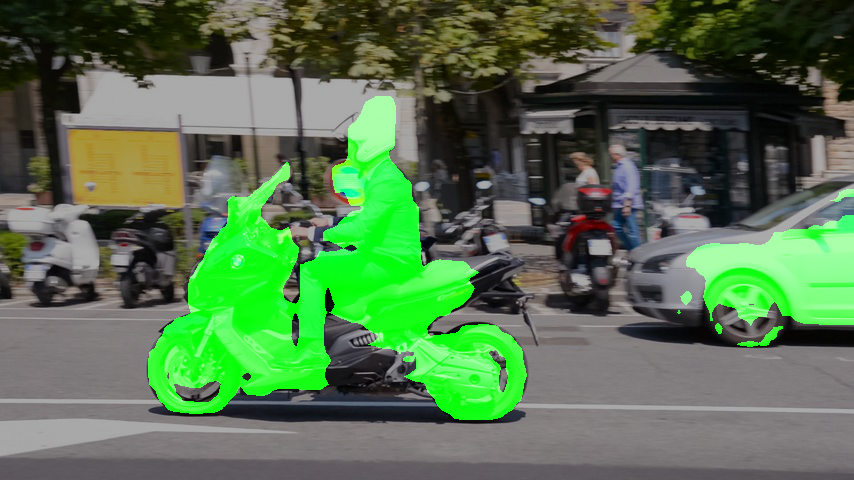}
\end{minipage}

\begin{minipage}{0.12\textwidth}
\begin{center}
\vfill
\small
adaptation\\
targets
\vfill
\end{center}
\end{minipage}
\begin{minipage}{0.87\textwidth}
\includegraphics[scale=\myscale]{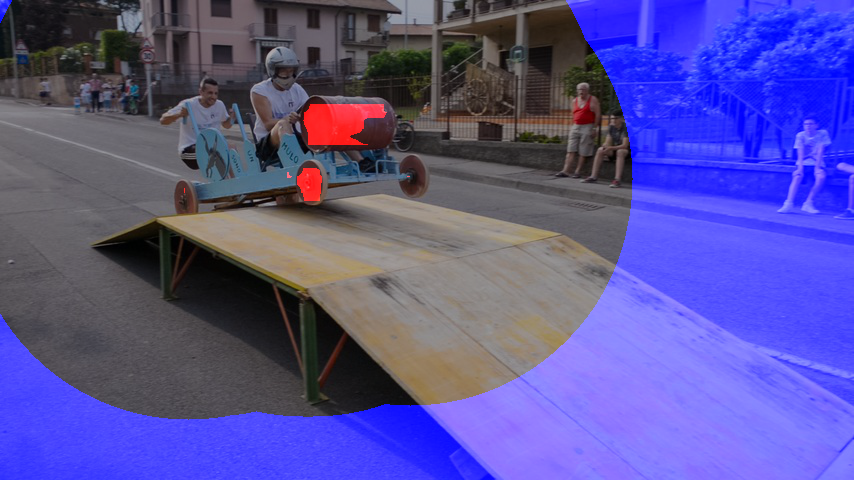}
\includegraphics[scale=\myscale]{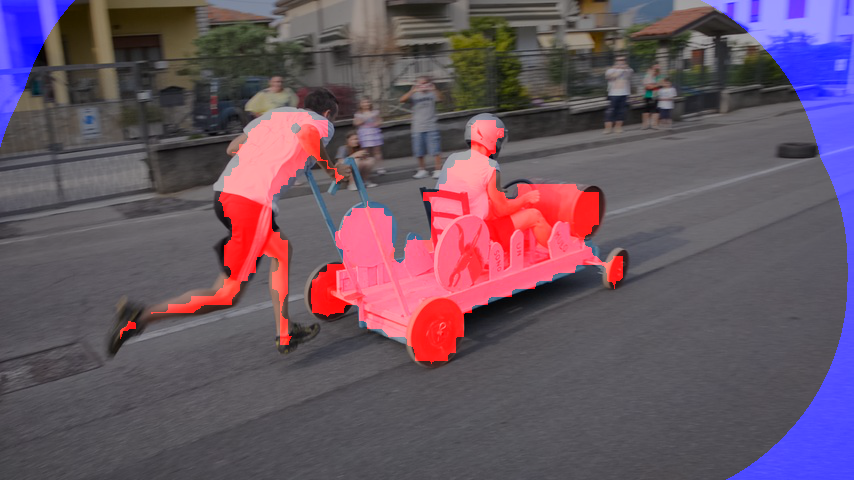}
\hspace{0.1cm}
\includegraphics[scale=\myscale]{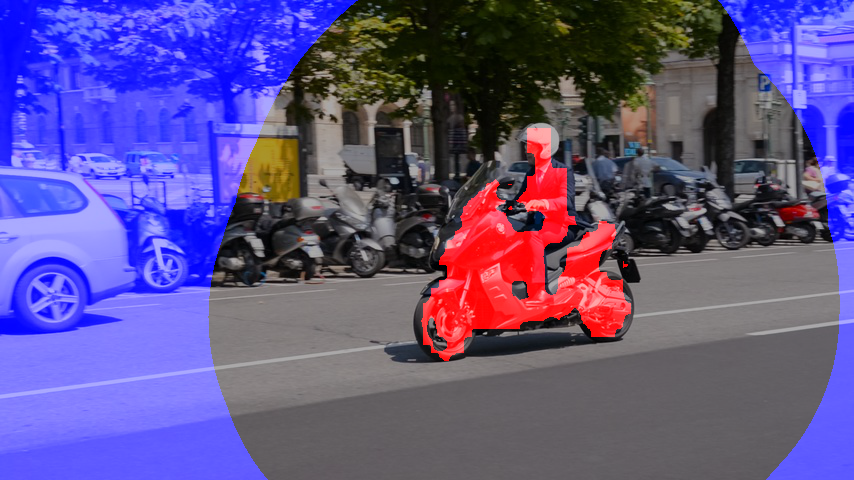}
\includegraphics[scale=\myscale]{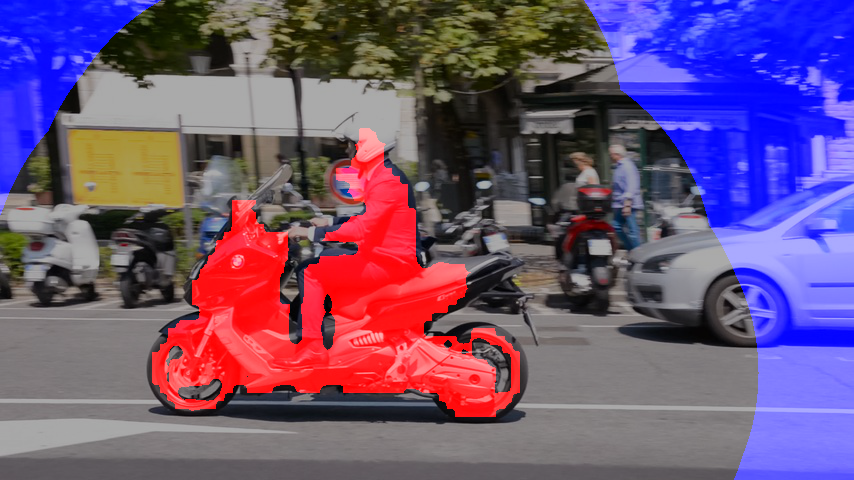}
\end{minipage}

\begin{minipage}{0.12\textwidth}
\begin{center}
\vfill
\small
online\\
adapted
\vfill
\end{center}
\end{minipage}
\begin{minipage}{0.87\textwidth}
\includegraphics[scale=\myscale]{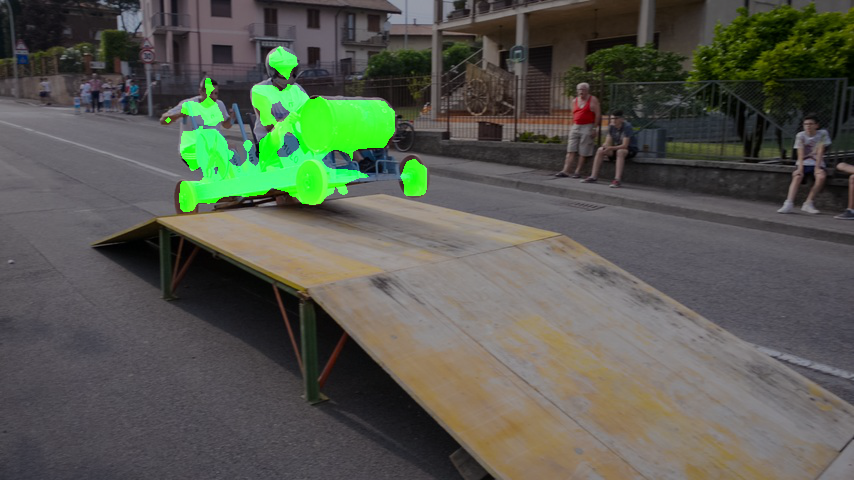}
\includegraphics[scale=\myscale]{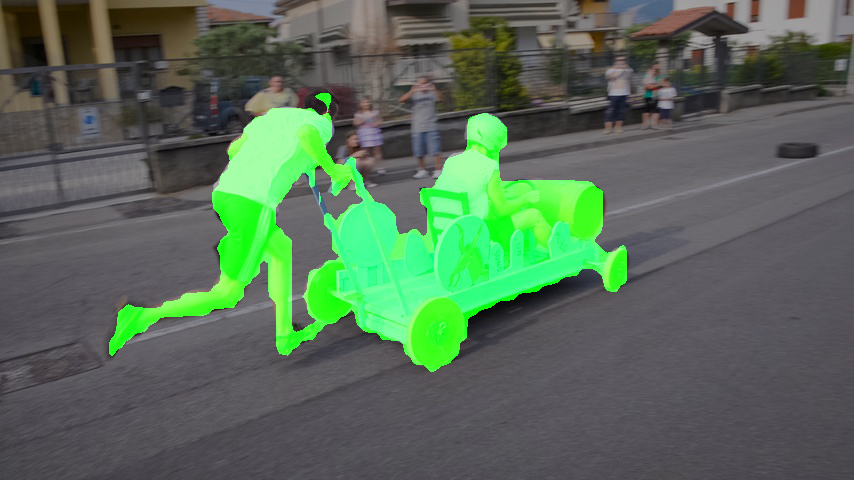}
\hspace{0.1cm}
\includegraphics[scale=\myscale]{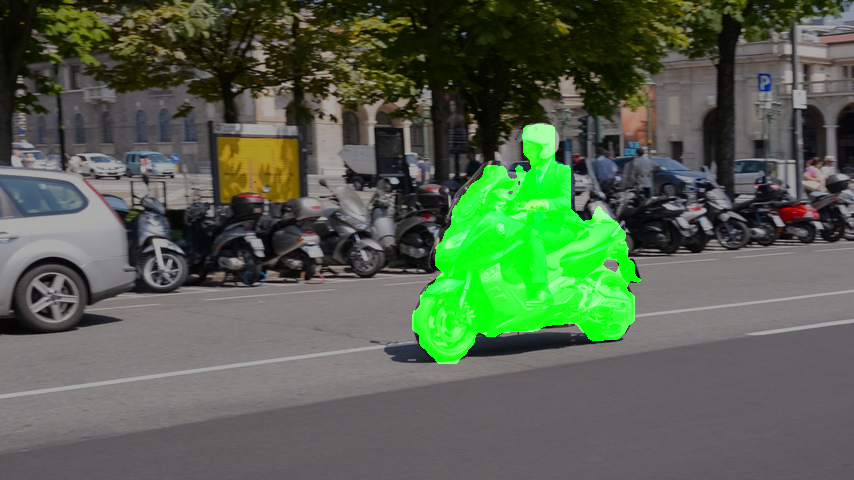}
\includegraphics[scale=\myscale]{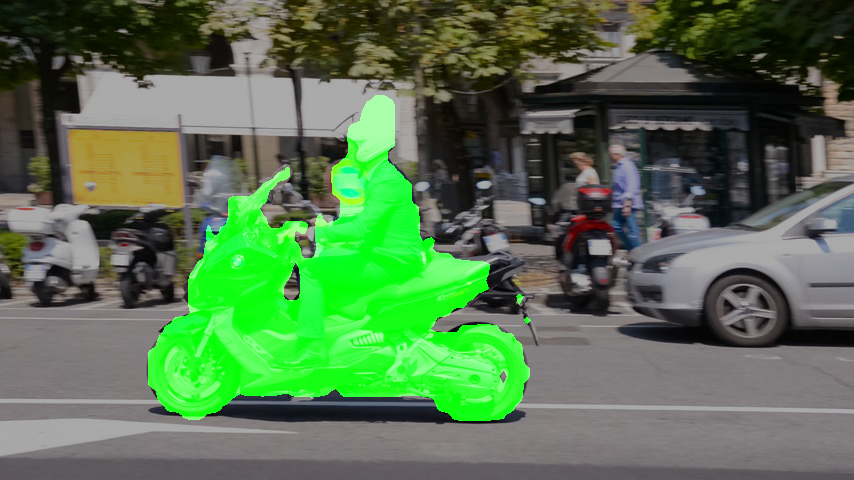}
\end{minipage}

\begin{minipage}{0.12\textwidth}
\begin{center}
\vfill
\small
ground\\
truth
\vfill
\end{center}
\end{minipage}
\begin{minipage}{0.87\textwidth}
\includegraphics[scale=\myscale]{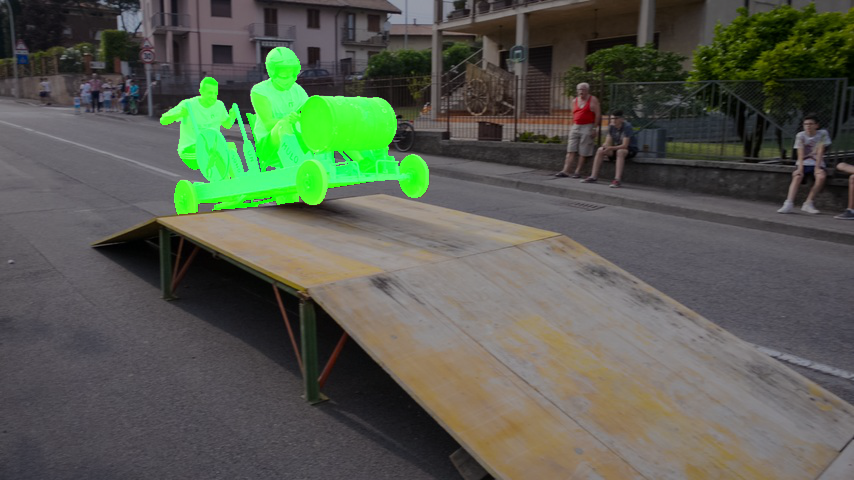}
\includegraphics[scale=\myscale]{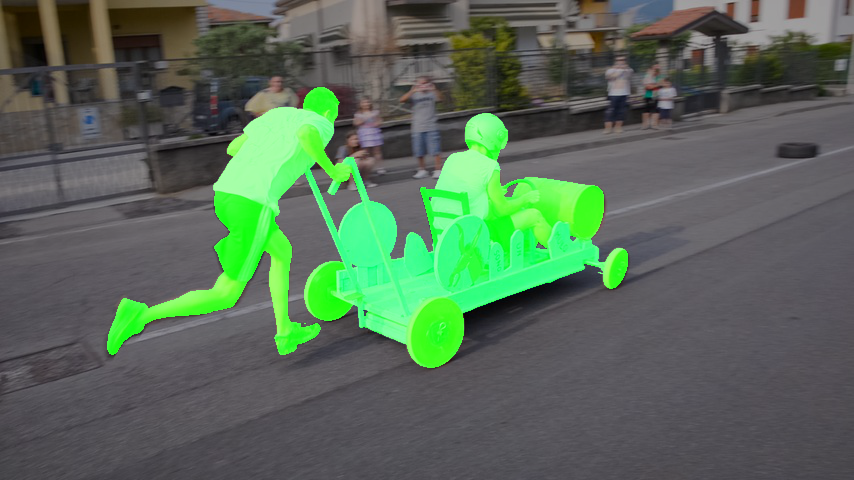}
\hspace{0.1cm}
\includegraphics[scale=\myscale]{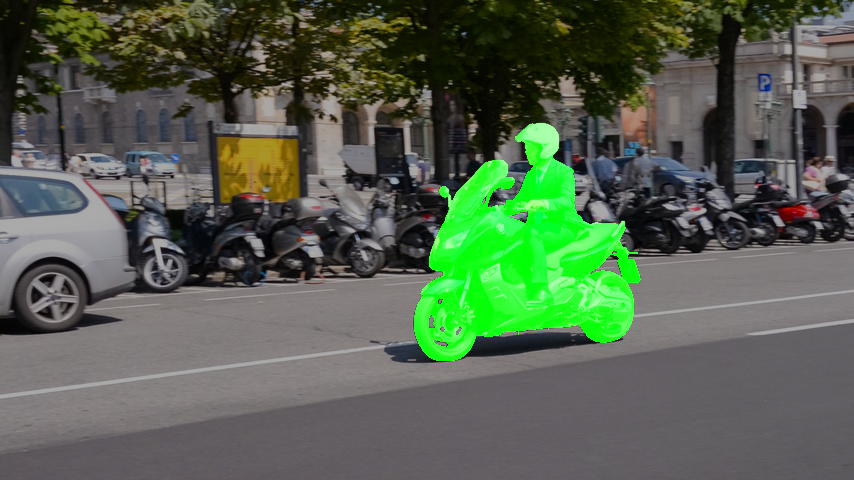}
\includegraphics[scale=\myscale]{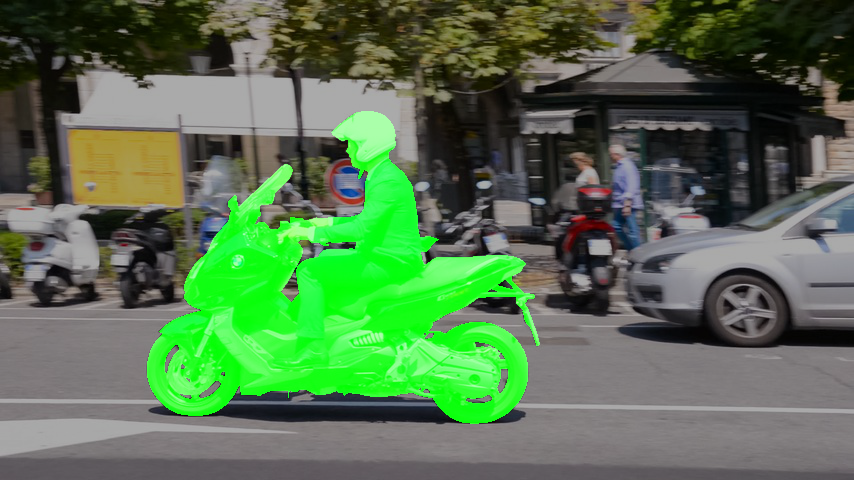}
\end{minipage}

\end{center}

\caption{\label{fig:teaser}Qualitative results on two sequences of the DAVIS validation set. The second row shows the pixels selected as positive (red) and negative (blue) training examples. It can be seen that after online adaptation, the network can deal better with changes in viewpoint (left) and new objects appearing in the scene (the car in the right sequence).}
\end{figure}

Visual object tracking is a fundamental problem in computer vision with many applications including video editing, autonomous cars, and robotics. Recently, there has been a trend to move from bounding box level to pixel level tracking, mainly driven by the availability of new datasets, in particular DAVIS \cite{DAVIS2016}. In our work, we focus on semi-supervised video object segmentation (VOS), \ie the task of segmenting the pixels belonging to a generic object in the video using the ground truth pixel mask of the first frame.

Recently, deep learning based approaches, which often utilize large classification datasets for pretraining, have shown extremely good performance for VOS \cite{OSVOS, masktrack, lucidtracker, fusionseg} and the related tasks of single-object tracking \cite{mdnet, GOTURN, Bertinetto2016ECCV} and background modeling \cite{Babaee17arxiv, Braham16IWSSIP, Wang16PRL}. 
In particular, the one-shot video object segmentation (\methodname{OSVOS}) approach introduced by Caelles \etal \cite{OSVOS}, has shown very promising results for VOS. This approach fine-tunes a pretrained convolutional neural network on the first frame of the target video. However, since at test time \methodname{OSVOS} only learns from the first frame of the sequence, it is not able to adapt to large changes in appearance, which might for example be caused by drastic changes in viewpoint.

While online adaptation has been used with success for bounding box level tracking (\eg \cite{TLD, boostingandvision, mdnet, Wang13NIPS, Li16TIP}), its use for VOS \cite{Papoutsakis13IVC, Ellis12ACCV, Bai09TOG, Bai10ECCV} has received less attention, especially in the context of deep learning. We thus propose Online Adaptive Video Object Segmentation (\methodname{OnAVOS}), which updates a convolutional neural network based on online-selected training examples. In order to avoid drift, we carefully select training examples by choosing pixels for which the network is very certain that they belong to the object of interest as positive examples, and pixels which are far away from the last assumed pixel mask as negative examples (see Fig. \ref{fig:teaser}, second row). We further show that naively performing online updates on every frame quickly leads to drift, which manifests in strongly degraded performance. As a countermeasure, we propose to mix in the first frame (for which the ground truth pixel mask is known) as additional training example during online updates.

Our contributions are the following: We introduce \methodname{OnAVOS}, which uses online updates to adapt to changes in appearance. Furthermore, we adopt a more recent network architecture and an additional objectness pretraining step \cite{pixelobjectness, fusionseg} and demonstrate their effectiveness for the semi-supervised setup. We further show that \methodname{OnAVOS} significantly improves the state of the art on two datasets.

\vspace{-1pt}
\section{Related Work}

\PARbegin{Video Object Segmentation.}
A common approach of many classical video object segmentation (VOS) methods is to reduce the granularity of the input space, \eg by using superpixels \cite{temporal_superpixels, eff_hier_graph}, patches  \cite{seamseg, jumpcut}, or object proposals \cite{fcproposals}. While these methods significantly reduce the complexity of subsequent optimization steps, they can introduce unrecoverable errors early in the pipeline. The obtained intermediate representations (or directly the pixels \cite{bilateralvidseg}) are then used for either a global optimization over the whole video \cite{bilateralvidseg, fcproposals}, over parts of it \cite{eff_hier_graph}, or using only the current and the preceding frame \cite{temporal_superpixels, jumpcut, seamseg}.

Recently, neural network based approaches \cite{OSVOS, masktrack, lucidtracker, fusionseg} including \methodname{OSVOS} \cite{OSVOS} have become the state of the art for VOS. Since \methodname{OnAVOS} is built on top of \methodname{OSVOS}, we include a detailed description in Section \ref{sec:OSVOS}. 
While \methodname{OSVOS} handles every video frame in isolation, we expect that incorporating temporal context should be helpful. As a step in this direction, Perazzi \etal \cite{masktrack} propose the \methodname{MaskTrack} method, in which the estimated segmentation mask from the last frame is used as an additional input channel to the neural network, enabling it to use temporal context. Jampani \etal \cite{videopropnetworks} propose a video propagation network (\methodname{VPN}) which applies learned bilateral filtering operations to propagate information across video frames. Furthermore, optical flow has been used as an additional temporal cue in conjunction with deep learning in the semi-supervised \cite{masktrack, lucidtracker} and unsupervised setting \cite{motionpatterns}, in which the ground truth for the first frame is not available. In our work, we focus on including context information implicitly by adapting the network online, \ie we store temporal context information in the adapted weights of the network.

Recently, Jain \etal \cite{pixelobjectness} proposed to train a convolutional neural network for pixel objectness, \ie for deciding for each pixel whether it belongs to an object-like region. In another paper, Jain \etal \cite{fusionseg} showed that using pixel objectness is helpful in the unsupervised VOS setting. We adopt pixel objectness as a pretraining step for the semi-supervised setting based on the one-shot approach.

The current best result on DAVIS is obtained by \methodname{LucidTracker} from Khoreva \etal \cite{lucidtracker}, which extends \methodname{MaskTrack} by an elaborate data augmentation method, which creates a large number of training examples from the first annotated frames and reduces the dependence on large datasets for pretraining. Our experiments show that our approach achieves better performance using only conventional data augmentation methods.

\PAR{Online Adaptation.}
For bounding box level tracking, Kalal \etal \cite{TLD} introduced the \methodname{Tracking-Learning-Detection (TLD)} framework, which tries to detect errors of the used object detector and to update the detector online to avoid these errors in the future.
Grabner and Bischof \cite{boostingandvision} used an online version of AdaBoost \cite{adaboost} for multiple computer vision tasks including tracking.
Nam and Han \cite{mdnet} proposed a \methodname{Multi-Domain Network} (\methodname{MDNet}) for bounding box level tracking. \methodname{MDNet} trains a separate domain-specific output layer for each training sequence and at test time initializes a new output layer, which is updated online together with two fully-connected layers. To this end, training examples are randomly sampled close to the current assumed object position, and are used as either positive or negative targets, based on their classification scores.
This scheme of sampling training examples online has some similarities to our approach. However, our method works on the pixel level instead of the bounding box level and, in order to avoid drift, we take special care to only select training examples online for which we are very certain that they are positive or negative examples.
For VOS, online adaptation is less well explored; mainly classical methods like online-updated color and/or shape models \cite{Bai10ECCV, Bai09TOG, Papoutsakis13IVC} and online random forests \cite{Ellis12ACCV} have been proposed.
\PAR{Fully Convolutional Networks for Semantic Segmentation.}
Fully Convolutional Networks (FCNs) for semantic segmentation have been introduced by Long \etal \cite{FCN}. The main idea is to repurpose a network initially designed for classification for semantic segmentation by replacing the fully-connected layers with $1\times1$ convolutions, and by introducing skip connections which help capture higher resolution details. Variants of this approach have since been widely adopted for semantic segmentation with great success (\eg ResNets by He \etal \cite{resnet}).

Recently, Wu \etal \cite{widerordeeper} introduced a ResNet variant with fewer but wider layers than the original ResNet architectures \cite{resnet} and a simple approach for segmentation, which avoids some of the subsampling steps by replacing them by dilated convolutions \cite{Yu16ICLR} and which does not use any skip connections. Despite the simplicity of their architecture for segmentation, they obtained outstanding results across multiple classification and semantic segmentation datasets, which motivates us to adopt their architecture.

\section{One-Shot Video Object Segmentation}
\label{sec:OSVOS}

\begin{figure}
\includegraphics[scale=0.6]{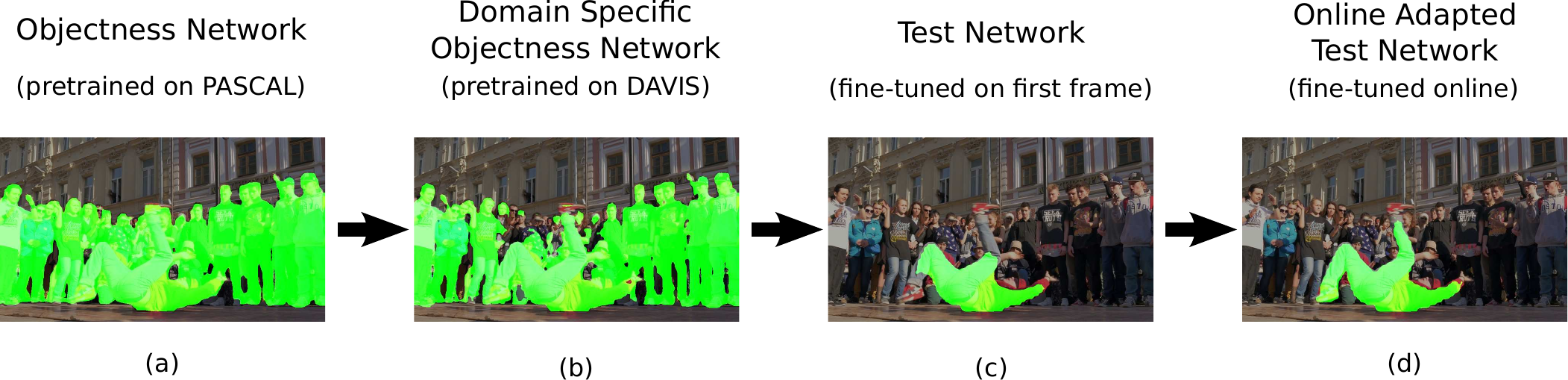}
\caption{\label{fig:pipeline}The pipeline of \methodname{OnAVOS}. Starting from pretrained weights, the network is first pretrained for objectness on PASCAL (a). Afterwards we pretrain on DAVIS to incorporate domain specific information (b). During test time, we fine-tune on the first frame, to obtain the test network (c). On the following frames, the network is then fine-tuned online to adapt to the changes in appearance (d). 
}
\end{figure}

\methodname{OnAVOS} (see Fig.~\ref{fig:pipeline} for an overview) builds upon the recently introduced one-shot video object segmentation (\methodname{OSVOS}) approach \cite{OSVOS}, but introduces pretraining for pixel objectness \cite{pixelobjectness} as a new component, adopts a more recent network architecture, and incorporates a novel online adaptation scheme, which is described in detail in Section \ref{sec:online}.

\PAR{Base Network.}
The first step of \methodname{OnAVOS} is to pretrain a base network on large datasets (\eg ImageNet \cite{imagenet} for image classification) in order to learn a powerful representation of objects, which can later be used as a starting point for the video object segmentation (VOS) task.

\PAR{Objectness Network.}
In a second step, the network is further pretrained for pixel objectness \cite{pixelobjectness} using a binary cross-entropy loss. In order to obtain targets for foreground and background, we use the PASCAL \cite{pascalVOC} dataset and map all 20 annotated classes to foreground and all other image regions are treated as background. As demonstrated by Jain \etal \cite{fusionseg}, the resulting objectness network alone already performs well on DAVIS, but here we use objectness only as a pretraining step.

\PAR{Domain Specific Objectness Network.}
The objectness network was trained on the \mbox{PASCAL} dataset. However, the target dataset on which the VOS should be performed may exhibit different characteristics, \eg a higher resolution and less noise in the case of DAVIS. Hence, we fine-tune the objectness network using the DAVIS training data and obtain a domain specific objectness network.
The DAVIS annotations do not directly correspond to objectness, as usually only one object out of possibly multiple is annotated. However, we argue that the learned task here is still similar to general objectness, since in most sequences of DAVIS the number of visible objects is relatively low and the object of interest is usually relatively large and salient.
Note that \methodname{OSVOS} trained the base network directly on DAVIS without objectness pretraining on PASCAL. Our experiments show that both steps are complementary.

\PAR{Test Network.}
After the preceding pretraining steps, the network has learned a domain specific notion of objectness, but during test time, it does not know yet which of the possibly multiple objects of the target sequence it should segment. Hence, we fine-tune the pretrained network on the ground truth mask of the first frame, which provides it with the identity and specific appearance of the object of interest and allows it to learn to ignore the background.
This one-shot step has been shown to be very effective for VOS \cite{OSVOS}, which we also confirm in our experiments. However, the first frame does not provide enough information for the network to adapt to drastic changes in appearance or viewpoint. In these cases, our online adaptation approach (see Section \ref{sec:online}) is needed.

\PAR{Network Architecture.}
While \methodname{OSVOS} used a variant of the well-known VGG network \cite{VGG}, we choose to adopt a more recent network architecture which incorporates residual connections. In particular, we adopt model A from Wu \etal \cite{widerordeeper}, which is a very wide ResNet \cite{resnet} variant with 38 hidden layers and roughly 124 million parameters. The approach for segmentation is very simple, as no upsampling mechanism or skip connections are used. Instead, downsampling by a factor of two using strided convolutions is performed only three times. This leads to a loss of resolution by a factor of eight in each dimension, following which the receptive field is increased using dilated convolutions \cite{Yu16ICLR} at no additional loss of resolution. Despite its simplicity, this architecture has shown excellent results both for classification (ImageNet) and segmentation (PASCAL) tasks \cite{widerordeeper}.
When applying it for segmentation, we bilinearly upsample the pixelwise posterior probabilities to the initial resolution before thresholding with $0.5$.

We use the weights provided by Wu \etal \cite{widerordeeper}, which were obtained by pretraining on ImageNet \cite{imagenet}, Microsoft COCO \cite{coco}, and PASCAL \cite{pascalVOC}, as a very strong initialization for the base network. We then replace the output layer with a two-class softmax. 
As loss function, we use the bootstrapped cross-entropy loss function \cite{bootstrappedCE}, which takes the average over the cross-entropy loss values only over a fraction of the hardest pixels, \ie pixels which are predicted worst by the network, instead of all pixels. This loss function has been shown to work well for unbalanced class distributions, which also commonly occur for VOS due to the dominant background class. In all our experiments, we use a fraction of 25\% of the hardest pixels and optimize this loss using the Adam optimizer \cite{adam}. In our evaluations, we separate the effect of the network architecture from the effect of the algorithmic improvements.

\section{Online Adaptation}
\label{sec:online}

\begin{algorithm}[t]
\renewcommand{\algorithmicrequire}{\textbf{Input:}}
\renewcommand{\algorithmicensure}{\textbf{Output:}}
\footnotesize
\caption{\label{alg:online}Online Adaptive Video Object Segmentation (\methodname{OnAVOS})}
\begin{multicols}{2}
\begin{algorithmic}[1]
\Require Objectness network $\mathcal{N}$, positive threshold $\alpha$, distance threshold $d$, total online steps $n_{online}$, current frame steps $n_{curr}$
\State Fine-tune $\mathcal{N}$ for 50 steps on $frame(1)$
\State $lastmask \gets ground\_truth(1)$
\For{$t=2\dots T$}
  \State $lastmask \gets erosion(lastmask)$
  \State $dtransform \gets distance\_transform(lastmask)$
  \State $negatives \gets dtransform > d$
  \State $posteriors \gets forward(\mathcal{N}, frame(t))$
  \State $positives \gets (posteriors > \alpha) \setminus negatives$
  \vfill\null
  \If{$lastmask \neq \emptyset$}
    \State interleaved:
    \State \quad Fine-tune $\mathcal{N}$ for $n_{curr}$ steps on $frame(t)$ \par
      \qquad \quad using $positives$ and $negatives$
    \State \quad Fine-tune $\mathcal{N}$ for $n_{online} - n_{curr}$ steps on \par
      \qquad \quad $frame(1)$ using $ground\_truth(1)$
  \EndIf
  \State $posteriors \gets forward(\mathcal{N}, frame(t))$
  \State $lastmask \gets (posteriors > 0.5) \setminus negatives$
  \State Output $lastmask$ for frame $t$
\EndFor
\end{algorithmic}
\end{multicols}
\vspace{-0.5cm}
\end{algorithm}

Since the appearance of the object of interest changes over time and new background objects can appear, we introduce an online adaptation scheme to adapt to these changes (see Algorithm \ref{alg:online}). New objects entering the scene are especially problematic when pretraining for objectness, since they were never used as negative training examples and are thus assigned a high probability (see Fig.~\ref{fig:teaser} (right) for an example).

The basic idea of our online adaptation scheme is to use pixels with very confident predictions as training examples. We select the pixels for which the predicted foreground probability exceeds a certain threshold $\alpha$ as positive examples. One could argue that using these pixels as positive examples is useless, since the network already gives very confident predictions for them. However, it is important that the adaptation retains a memory of the positive class in order to create a counterweight to the many negative examples being added. In our experiments, leaving out this step resulted in holes in the foreground mask.

We initially selected negative training examples in the same way, \ie using pixels with a very low foreground probability. However, this led to degraded performance, probably, because during large appearance changes, false negative pixels will be selected as negative training examples, effectively destroying all chances to adapt to these changes. We thus select negative training examples in a different way, based on the assumption that the movement between two frames is small. The idea is to select all pixels which are very far away from the last predicted object mask. In order to deal with noise, the last mask can first be shrunk by an erosion operation. For our experiments, we use a square structural element with size $15$, but we found that the exact value of this parameter is not critical.
Afterwards, we compute a distance transform, which for each pixel provides the Euclidean distance to the closest foreground pixel of the mask. Finally, we apply a threshold $d$ and treat all pixels with a distance larger than $d$ as negative examples.

Pixels which are neither marked as positive nor as negative examples are assigned a ``don't care'' label and are ignored during the online updates. 
We can now fine-tune the network on the current frame, since every pixel has a label for training. However, in practice, we found that naively fine-tuning using the obtained training examples quickly leads to drift.
To circumvent this problem, we propose to mix in the first frame as additional training examples during the online updates, since for the first frame the ground truth is available. 
We found that in order to obtain good results, the first frame should be sampled more often than the current frame, \ie during online adaptation we perform a total of $n_{online}$ update steps per frame, of which only $n_{curr}$ are performed on the current frame, and the rest is performed on the first frame. Additionally, we reduce the weight of the loss for the current frame by a factor $\beta$ (\eg $\beta \approx 0.05$). A value of $0.05$ might seem surprisingly small, but one has to keep in mind that the first frame is used very often for updates, quickly leading to smaller gradients, while the current frame is only selected a few times.

During online adaptation, the negative training examples are selected based on the mask of the preceding frame. Hence, it can happen that a pixel is selected as a negative example and that it is predicted as foreground at the same time. We call such pixels hard negatives. A common case  in which hard negatives occur is when a previously unseen object enters the scene far away from the object of interest (see Fig.~\ref{fig:teaser} (right)), which will then usually be detected as foreground by the network.
We found it helpful to remove hard negatives from the foreground mask which is used in the next frame to determine negative training examples. This step allows selecting the hard negatives in the next frame again as negative examples.
Additionally, we tried to adapt the network more strongly to hard negatives by increasing the number of update steps and/or the loss scale for the current frame in the presence of hard negatives. However, this did not improve the results further.

In addition to the previously described steps, we propose a simple heuristic which makes our method more robust against difficulties like occlusion: If (after the optional erosion) nothing is left of the last assumed foreground mask, we assume that the object of interest is lost and do not apply any online updates until the network again finds a non-empty foreground mask.

\section{Experiments}
\PARbegin{Datasets.}
For objectness pretraining (\cf Section \ref{sec:OSVOS}), we used the 1,464 training images of the PASCAL VOC 2012 dataset \cite{pascalVOC} plus the additional annotations provided by Hariharan \etal \cite{PASCALaug}, leading to a total of 10,582 training images with 20 classes, which we all mapped to a single foreground class. For video object segmentation (VOS), we conducted most experiments on the recently introduced DAVIS dataset \cite{DAVIS2016}, which consists of 50 short full-HD video sequences, from which 30 are taken for training and 20 for validation. Consistent with most prior work, we conduct all experiments on the subsampled version with a resolution of $854 \times 480$ pixels. In order to show that our method generalizes, we also conducted experiments on the YouTube-Objects \cite{YoutubeObjectsOriginal, YoutubeObjectsSegmentation} dataset for VOS, consisting of 126 sequences.


\PAR{Experimental Setup.}
We pretrain on PASCAL and DAVIS, for 10 epochs each. For the baseline one-shot approach, we found 50 update steps on the first frame with a learning rate of $3 \cdot 10^{-6}$ to work well. For simplicity, we used a mini-batch size of only one image.
Since DAVIS only has a training and a validation set, we tuned all hyperparameters on the training set of 30 sequences using three-fold cross validation, \ie 20 training sequences are used for training and 10 for validation for each fold. As is standard practice, we augmented the training data by random flipping, scaling with a factor uniformly sampled from $[0.7, 1.3]$, and gamma augmentations \cite{FRRN}. 

For evaluation, we used the Jaccard index, \ie the mean intersection-over-union (mIoU) between the predicted foreground masks and the ground truth masks. Results for additional evaluation measures suggested by Perazzi \etal \cite{DAVIS2016} are shown in the supplementary material. We noticed that, especially for fine-tuning on the first frame, the random augmentations introduce non-negligible variations in the results. Hence, for these experiments, we conducted three runs and report mean and standard deviation values. All experiments were performed with our TensorFlow \cite{tensorflow2015-whitepaper} based implementation, which we will make available together with pretrained models at \url{https://www.vision.rwth-aachen.de/software/OnAVOS}.

\subsection{Baseline Systems}

\begin{table}
\small
\begin{center}
\iftrue
\begin{tabular}{|c|c|c|c|}
\hline
PASCAL & DAVIS & First frame & mIoU {[}\%{]}\tabularnewline
\hline
\hline
\checkmark & \checkmark & \checkmark & $\pmb{80.3}\pm0.4$\tabularnewline
\hline
 & \checkmark & \checkmark & $78.0\pm0.1$\tabularnewline
\hline
\checkmark &  & \checkmark & $77.6\pm0.4$\tabularnewline
\hline
\checkmark & \checkmark &  & $72.7$\tabularnewline
\hline
\checkmark &  &  & $65.3$\tabularnewline
\hline
 & \checkmark &  & $71.0$\tabularnewline
\hline
 &  & \checkmark & $65.2\pm1.0$\tabularnewline
\hline
\end{tabular}
\else
\begin{tabular}{|c||c|c|c|c|c|c|c|}
\hline
PASCAL & \checkmark &  & \checkmark & \checkmark & \checkmark &  & \tabularnewline
\hline
DAVIS & \checkmark & \checkmark &  & \checkmark &  & \checkmark & \tabularnewline
\hline
First frame & \checkmark & \checkmark & \checkmark &  &  &  & \checkmark\tabularnewline
\hline
\hline
mIoU {[}\%{]} & $80.3\pm0.4$ & $78.0\pm0.1$ & $77.6\pm0.4$ & $72.7$ & $65.3$ & $71.0$ & $65.2\pm1.0$\tabularnewline
\hline
\end{tabular}
\fi
\end{center}
\caption{\label{tab:ablations}Effect of (pre-)training steps on the DAVIS validation set. As can be seen, each of the three training steps 
are useful. The objectness pretraining step on PASCAL significantly improves the results.}
\end{table}

\PARbegin{Effect of Pretraining Steps.}
Starting from the base network (\cf Section \ref{sec:OSVOS}) our full baseline system (\ie without adaptation) includes a first pretraining step on PASCAL for objectness, then on the training sequences of DAVIS, and finally a one-shot fine-tuning on the first frame. Each of these three steps can be enabled or disabled individually. Table \ref{tab:ablations} shows the results on DAVIS for all resulting combinations. As can be seen, each of these steps is useful since removing any step always deteriorates the results.

The base network was trained for a different task than binary segmentation and thus a new output layer needs to be learned at the same time as fine-tuning the rest of the network. Without pretraining on either PASCAL or DAVIS, the randomly initialized output layer is learned only from the first frame of the target sequence, which leads to a largely degraded performance of only 65.2\% mIoU. However, when either PASCAL or DAVIS is used for pretraining, the result is greatly improved to 77.6\% mIoU and 78.0\% mIoU, respectively. While both results are very similar, it can be seen that PASCAL and DAVIS do provide complementary information, since using both datasets together further improves the result to 80.3\%. 
We argue that the relatively large PASCAL dataset is useful for learning general objectness, while the limited amount of DAVIS data is useful to adapt to the characteristics (\eg relatively high image quality) of the data of DAVIS, which provides an advantage for evaluating on DAVIS sequences.

Interestingly, even without looking at the segmentation mask of the first frame, \ie in the unsupervised setup, we already obtain a result of 72.7\% mIoU; slightly better than the current best unsupervised method FusionSeg \cite{fusionseg}, which obtains 70.7\% mIoU on the DAVIS validation set\footnote{In FusionSeg \cite{fusionseg}, the result for all sequences including the training set is reported, but here we calculated the average only over the validation sequences for better comparability} using objectness and optical flow as an additional cue.

\PAR{Comparison to \methodname{OSVOS}.}
Without including their boundary snapping post-processing step, \methodname{OSVOS} achieves a result of 77.4\% mIoU on DAVIS. Our system without objectness pretraining on PASCAL is directly comparable to this result and achieves 78.0\% mIoU. We attribute this moderate improvement to the more recent network architecture which we adopted. Including PASCAL for objectness pretraining improves this result by further 2.3\% to 80.3\%.

\subsection{Online Adaptation}

%
%

\PARbegin{Hyperparameter Study.}
As described in Section \ref{sec:online}, \methodname{OnAVOS} involves relatively many hyperparameters. After some coarse manual tuning on the DAVIS training set, we found $\alpha=0.97$, $\beta=0.05$, $d=220$, $n_{online}=15$, $n_{curr}=3$ to work well. While the initial $50$ update steps on the first frame are performed with a learning rate of $3\cdot 10^{-6}$, it proved useful to use a different learning rate $\lambda=10^{-5}$ for the online updates on the current and the first frame.
Starting from these values as the operating point, we conducted a more detailed study by changing one hyperparameter at a time, while keeping the others constant. We found that \methodname{OnAVOS} is not very sensitive to the choice of most hyperparameters and each configuration we tried performed better than the non-adapted baseline and we achieved only small improvements compared to the operating point (detailed plots are shown in the supplementary material). To avoid overfitting to the small DAVIS training set, we kept the values from the operating point for all further experiments.

\PAR{Ablation Study.}
Table \ref{tab:online} shows the results of the proposed online adaptation scheme and multiple variants, where parts of the algorithm are disabled, on the DAVIS validation set. Using the full method, we obtain an mIoU score of $82.8\%$. When disabling all adaptation steps, the performance significantly degrades to $80.3\%$, which demonstrates the effectiveness of the online adaptation method.
The table further shows that negative training examples are more important than positive ones. 
If we do not mix in the first frame during online updates, the result is significantly degraded to 69.1\% due to drift.

\PAR{Timing Information.}
For the initial fine-tuning stage on the first frame, we used 50 update steps. Including the time for the forward pass for all further frames, this leads to a total runtime of around 90 seconds per sequence (corresponding to roughly 1.3 seconds per frame) of the DAVIS validation set using an NVIDIA Titan X (Pascal) GPU. When using online adaptation with $n_{online}=15$, the runtime increases to around 15 minutes per sequence (corresponding to roughly 13 seconds per frame). However, our hyperparameter analysis revealed that this runtime can be significantly decreased by reducing $n_{online}$ without much loss of accuracy. Note that for best results, \methodname{OSVOS} used a higher number of update steps on the first frame and needs about 10 minutes per sequence (corresponding to roughly 9 seconds per frame). 

\begin{table}
\small
\begin{center}
\begin{tabular}{|c|c|}
\hline
Method & mIoU {[}\%{]}\tabularnewline
\hline
\hline
No adaptation & $80.3\pm0.4$\tabularnewline
Full adaptation & $\pmb{82.8}\pm0.5$\tabularnewline
Only negatives & $82.4\pm0.3$\tabularnewline
Only positives & $81.6\pm0.3$\tabularnewline
No first frame during online adaptation & $69.1\pm0.2$\tabularnewline
\hline
\end{tabular}
\end{center}
\caption{\label{tab:online}Online adaptation ablation experiments on the DAVIS validation set. As can be seen, mixing in the first frame during online updates is essential, and negative examples are more important than positive ones.}
\end{table}

\subsection{Comparison to State of the Art}

\begin{table}
\small
\begin{center}
\begin{tabular}{|c|c|c|}
\hline
\multirow{2}{*}{Method} & DAVIS & YouTube-Objects\tabularnewline
 & mIoU {[}\%{]} & mIoU {[}\%{]}\tabularnewline
\hline
\hline
\methodname{OnAVOS} (ours), no adaptation & $80.3\pm0.4$ & $76.1\pm1.3$ \tabularnewline
+CRF & $81.7\pm0.5$ & $76.4\pm0.2$ \tabularnewline
+CRF +Test time augmentations & $81.7\pm0.2$ & $76.6\pm0.1$ \tabularnewline
\hline
\methodname{OnAVOS} (ours), online adaptation & $82.8\pm0.5$ & $76.8\pm0.1$ \tabularnewline
+CRF & $84.3\pm0.5$ & $77.2\pm0.2$\tabularnewline
+CRF +Test time augmentations & $\pmb{85.7}\pm0.6$ & $\pmb{77.4}\pm0.2$ \tabularnewline
\hline
\methodname{OSVOS} \cite{OSVOS} & $79.8$ & $72.5$\tabularnewline
\methodname{MaskTrack} \cite{masktrack} & $79.7$ & $72.6$\tabularnewline
\methodname{LucidTracker} \cite{lucidtracker} $^{\dagger}$ & $80.5$ & $\mathit{76.2}$\tabularnewline
\methodname{VPN} \cite{videopropnetworks} & $75.0$ & - \tabularnewline
\hline
\end{tabular}
\end{center}
\caption{\label{tab:SOTA}Comparison to the state of the art on the DAVIS validation set and the YouTube-Objects dataset. $\dagger$: Concurrent
work only published on arXiv. More results are shown in the supplementary material.}
\end{table}

Current state of the art methods use post-processing steps such as boundary snapping \cite{OSVOS}, or conditional random field (CRF) smoothing \cite{masktrack, lucidtracker} to improve the contours. In order to compare with them, we included per-frame post-processing using DenseCRF \cite{CRF}. This might be especially useful since our network only provides one output for each $8\times8$ pixel block. Additionally, we added data augmentations during test time. To this end, we created 10 variants of each test image by random flipping, zooming, and gamma augmentations, and averaged the posterior probabilities over all 10 images. 

In order to demonstrate the generalization ability of \methodname{OnAVOS} and since there is no separate training set for YouTube-Objects, we conducted our experiments on this dataset using the same hyperparameter values as for DAVIS, including the CRF parameters. Additionally, we omitted the pretraining step on DAVIS. 
Note that for YouTube-Objects, the evaluation protocols in prior publications sometimes differed by not including frames in which the object of interest is not present \cite{lucidtracker}. Here, we report results following the DAVIS evaluation protocol, \ie including these frames, consistent with Khoreva \etal \cite{lucidtracker}.

Table \ref{tab:SOTA} shows the effect of our post-processing steps and compares our results on DAVIS and YouTube-Objects to other methods. Note that the effect of the test time augmentations is stronger when combined with online adaptation. We argue that this is because in this case, the augmentations do not only directly improve the end result as a post-processing step, but they also deliver better adaptation targets. On DAVIS, we achieve an mIoU of $85.7\%$ which is, to the best of our knowledge significantly higher than any previously published result. Compared to \methodname{OSVOS}, this is an improvement of almost 6\%. 
On YouTube-Objects, we achieve an mIoU of $77.4\%$, which is also a significant improvement over the second best result obtained by \methodname{LucidTracker} with $76.2\%$.

\section{Conclusion}
In this work, we have proposed \methodname{OnAVOS}, which builds on the \methodname{OSVOS} approach. We have demonstrated that the inclusion of an objectness pretraining step and our online adaptation scheme for semi-supervised video object segmentation are highly effective. We have further shown that our online adaptation scheme is robust against choices of hyperparameters and generalizes to another dataset. We expect that, in the future, more methods will adopt adaptation schemes which make them more robust against large changes in appearance. For future work, we plan to explicitly incorporate temporal context information into our method.

\vspace{0.6cm}
\PARbegin{Acknowledgements.}
The work in this paper is funded by the EU project STRANDS (ICT-2011-600623) and the ERC Starting
Grant project CV-SUPER (ERC-2012-StG-307432). We would like to thank Istv\'{a}n S\'{a}r\'{a}ndi, J\"{o}rg St\"{u}ckler, Lucas Beyer, and Aljo\v{s}a O\v{s}ep for helpful discussions and proofreading.

\bibliography{abbrev_short,arxiv}

\clearpage
\appendix
\textcolor{bmv@sectioncolor}{\vspace{-1.2cm} \part*{Supplementary Material}}
\normalsize

\section{More Comprehensive Comparison to Other Methods}

Table \ref{tab:comparison} shows a more comprehensive comparison of our results to the results obtained by other methods.

\begin{table}[h]
\footnotesize
\begin{center}
\begin{tabular}{|c|c|c|}
\hline
\multirow{2}{*}{Method} & DAVIS & YouTube-Objects\tabularnewline
 & mIoU {[}\%{]} & mIoU {[}\%{]}\tabularnewline
\hline
\hline
\methodname{OnAVOS} (ours), no adaptation & $81.7\pm0.2$ & $76.6\pm0.1$ \tabularnewline
\methodname{OnAVOS} (ours), online adaptation & $\pmb{85.7}\pm0.6$ & $\pmb{77.4}\pm0.2$ \tabularnewline 
\hline
\methodname{OSVOS} \cite{OSVOS} & $79.8$ & $72.5$\tabularnewline
\methodname{MaskTrack} \cite{masktrack} & $79.7$ & $72.6$\tabularnewline
\methodname{LucidTracker} \cite{lucidtracker} $^{\dagger}$ & $80.5$ & $76.2$\tabularnewline
\methodname{VPN} \cite{videopropnetworks} & $75.0$ & - \tabularnewline
\methodname{FCP} \cite{fcproposals} & 63.1 & - \tabularnewline
\methodname{BVS} \cite{bilateralvidseg} & 66.5 & 59.7 \tabularnewline
\methodname{OFL} \cite{Tsai16CVPR} & 71.1 & 70.1 \tabularnewline
\methodname{STV} \cite{Wang17arxiv} & 73.6 & - \tabularnewline

\hline
\end{tabular}
\end{center}
\caption{\label{tab:comparison}Comparison to other methods on the DAVIS validation set and the YouTube-Objects dataset. Note that \methodname{MaskTrack} \cite{masktrack} and \methodname{LucidTracker} \cite{lucidtracker} report results on DAVIS for all sequences including the training set, but here we show their results for the validation set only. $\dagger$: Concurrent work only published on arXiv.}
\end{table}

\section{Additional Evaluation Measures for DAVIS}
Table \ref{tab:measures} shows a more detailed evaluation on the DAVIS validation set using the evaluation measures suggested by Perazzi \etal \cite{DAVIS2016}. The measures used here are the Jaccard index $\mathcal{J}$, defined as the mean intersection-over-union (mIoU) between the predicted foreground masks and the ground truth masks; the contour accuracy measure $\mathcal{F}$, which measures how well the segmentation boundaries agree; and the temporal stability measure $\mathcal{T}$, which measures the consistency of the predicted masks over time. For more details of these measures, we refer the interested reader to Perazzi \etal \cite{DAVIS2016}. Note that the results for additional measures for \methodname{LucidTracker} \cite{lucidtracker} are missing since they are only reported averaged over all 50 sequences of DAVIS and not on the validation set.

The table shows that each evaluation measure is significantly improved by the proposed online adaptation scheme. \methodname{OnAVOS} obtains the best mean results for all three measures. It is surprising that our result for the temporal stability $\mathcal{T}$ is better than the result by \methodname{MaskTrack} \cite{masktrack}, although in contrast to our method, they explicitly incorporate temporal context by propagating masks.

\begin{table}[h]
\footnotesize
\begin{center}
\begin{tabular}{|cc|c|c|c|c|c|}
\hline 
\multicolumn{2}{|c}{\multirow{2}{*}{Measure}} & \multicolumn{2}{|c|}{\methodname{OnAVOS} (ours)} & \multirow{2}{*}{\methodname{OSVOS} \cite{OSVOS}} & \multirow{2}{*}{\methodname{MaskTrack} \cite{masktrack}} & \multirow{2}{*}{\methodname{LucidTracker} \cite{lucidtracker}}\tabularnewline
\cline{3-4} 
 &  & Un-adapted & Adapted &  &  & \tabularnewline
\hline 
\hline 
\multirow{3}{*}{$\mathcal{J}$} & mean $\uparrow$ & $\mathit{81.7}\pm0.2$ & $\pmb{85.7}\pm0.6$ & $79.8$ & $79.7$ & $80.5$\tabularnewline
 & recall $\uparrow$ & $92.2\pm0.6$ & $\pmb{95.4}\pm0.8$ & $\mathit{93.6}$ & $93.1$ & - \tabularnewline
 & decay $\downarrow$ & $11.9\pm0.3$ & $\pmb{7.1}\pm1.7$ & $14.9$ & $\mathit{8.9}$ & - \tabularnewline
\hline 
\multirow{3}{*}{$\mathcal{F}$} & mean $\uparrow$ & $\mathit{81.1}\pm0.2$ & $\pmb{84.2}\pm0.8$ & $80.6$ & $75.4$ & - \tabularnewline
 & recall $\uparrow$ & $88.2\pm0.3$ & $\mathit{88.7}\pm1.3$ & $\pmb{92.6}$ & $87.1$ & - \tabularnewline
 & decay $\downarrow$ & $11.2\pm0.5$ & $\pmb{7.8}\pm1.8$ & $15.0$ & $\mathit{9.0}$ & - \tabularnewline
\hline 
$\mathcal{T}$ & mean $\downarrow$ & $27.3\pm2.2$ & $\pmb{18.5}\pm0.1$ & $37.6$ & $\mathit{21.8}$ & - \tabularnewline
\hline 
\end{tabular}
\end{center}

\caption{\label{tab:measures}Additional evaluation measures on the DAVIS validation set. Best and second best results are highlighted with bold and italic fonts, respectively.}
\end{table}

\section{Per-Sequence Results for DAVIS}

Table \ref{tab:perseq} shows mIoU results for each of the 20 sequences of the DAVIS validation set. On 18 out of 20 sequences, \methodname{OnAVOS} obtains either the best or the second best result.

\begin{table}[h!]
\footnotesize
\begin{center}
\begin{tabular}{|c|c|c|c|c|c|}
\hline 
\multirow{3}{*}{Sequence} & \multicolumn{5}{c|}{Method, mIoU {[}\%{]}}\tabularnewline
\cline{2-6} 
 & \multicolumn{2}{c|}{OnAVOS (ours) } & \multirow{2}{*}{\methodname{OSVOS} \cite{OSVOS}} & \multirow{2}{*}{\methodname{MaskTrack} \cite{masktrack}} & \multirow{2}{*}{\methodname{LucidTracker} \cite{lucidtracker}}\tabularnewline
\cline{2-3} 
 & Un-adapted & Adapted &  &  & \tabularnewline
\hline 
\hline 
blackswan & $\mathit{96.1}\pm0.1$ & $\pmb{96.2}\pm0.1$ & $94.2$ & $90.3$ & $95.0$\tabularnewline
bmx-trees & $48.2\pm0.8$ & $\mathit{57.0}\pm1.0$ & $55.5$ & $\pmb{57.5}$ & $55.0$\tabularnewline
breakdance & $62.6\pm4.2$ & $73.6\pm3.8$ & $70.8$ & $\mathit{76.1}$ & $\pmb{87.2}$\tabularnewline
camel & $84.6\pm0.1$ & $\mathit{85.5}\pm0.1$ & $85.1$ & $80.1$ & $\pmb{94.3}$\tabularnewline
car-roundabout & $86.5\pm0.2$ & $\pmb{97.5}\pm0.0$ & $95.3$ & $\mathit{96.0}$ & $\mathit{96.0}$\tabularnewline
car-shadow & $\mathit{94.1}\pm0.1$ & $\pmb{96.8}\pm0.1$ & $93.7$ & $93.5$ & $90.3$\tabularnewline
cows & $\pmb{95.4}\pm0.0$ & $\pmb{95.4}\pm0.0$ & $\mathit{94.6}$ & $88.2$ & $93.1$\tabularnewline
dance-twirl & $78.4\pm0.7$ & $\mathit{85.6}\pm1.0$ & $67.0$ & $84.4$ & $\pmb{88.6}$\tabularnewline
dog & $\pmb{95.6}\pm0.1$ & $\pmb{95.6}\pm0.1$ & $90.7$ & $90.8$ & $\mathit{95.0}$\tabularnewline
drift-chicane & $\mathit{87.4}\pm0.5$ & $\pmb{89.2}\pm0.2$ & $83.5$ & $86.2$ & $1.4$\tabularnewline
drift-straight & $\mathit{81.3}\pm5.6$ & $\pmb{93.7}\pm0.9$ & $67.6$ & $56.0$ & $79.9$\tabularnewline
goat & $\mathit{90.8}\pm0.1$ & $\pmb{91.4}\pm0.1$ & $88.0$ & $84.5$ & $88.9$\tabularnewline
horsejump-high & $\mathit{89.3}\pm0.3$ & $\pmb{90.1}\pm0.0$ & $78.0$ & $81.8$ & $87.1$\tabularnewline
kite-surf & $\pmb{70.1}\pm1.0$ & $\mathit{69.1}\pm0.1$ & $68.6$ & $60.0$ & $64.6$\tabularnewline
libby & $\mathit{87.1}\pm1.0$ & $\pmb{88.6}\pm0.1$ & $80.8$ & $77.5$ & $85.5$\tabularnewline
motocross-jump & $\pmb{89.7}\pm0.2$ & $70.4\pm11.9$ & $\mathit{81.6}$ & $68.3$ & $75.1$\tabularnewline
paragliding-launch & $\pmb{64.6}\pm0.1$ & $\mathit{64.3}\pm0.1$ & $62.5$ & $62.1$ & $63.7$\tabularnewline
parkour & $92.4\pm0.2$ & $\pmb{93.6}\pm0.0$ & $85.6$ & $88.2$ & $\mathit{93.2}$\tabularnewline
scooter-black & $64.8\pm7.1$ & $\pmb{91.3}\pm0.1$ & $71.1$ & $82.4$ & $\mathit{86.5}$\tabularnewline
soapbox & $74.0\pm4.6$ & $\mathit{89.8}\pm1.2$ & $81.2$ & $\mathit{89.9}$ & $\pmb{90.5}$\tabularnewline
\hline 
mean & $\mathit{81.7}\pm0.2$ & $\pmb{85.7}\pm0.6$ & $79.8$ & $79.7$ & $80.5$\tabularnewline
\hline 
\end{tabular}
\end{center}
\caption{\label{tab:perseq}Per-sequence results on the DAVIS validation set. Best and second best results are highlighted with bold and italic fonts, respectively.}

\end{table}



\section{Hyperparameter Study on DAVIS}\label{sec:hyperparams}

\begin{figure}[h!]
\centering
\subfigure[online learning rate $\lambda$]{\includegraphics[scale=0.23]{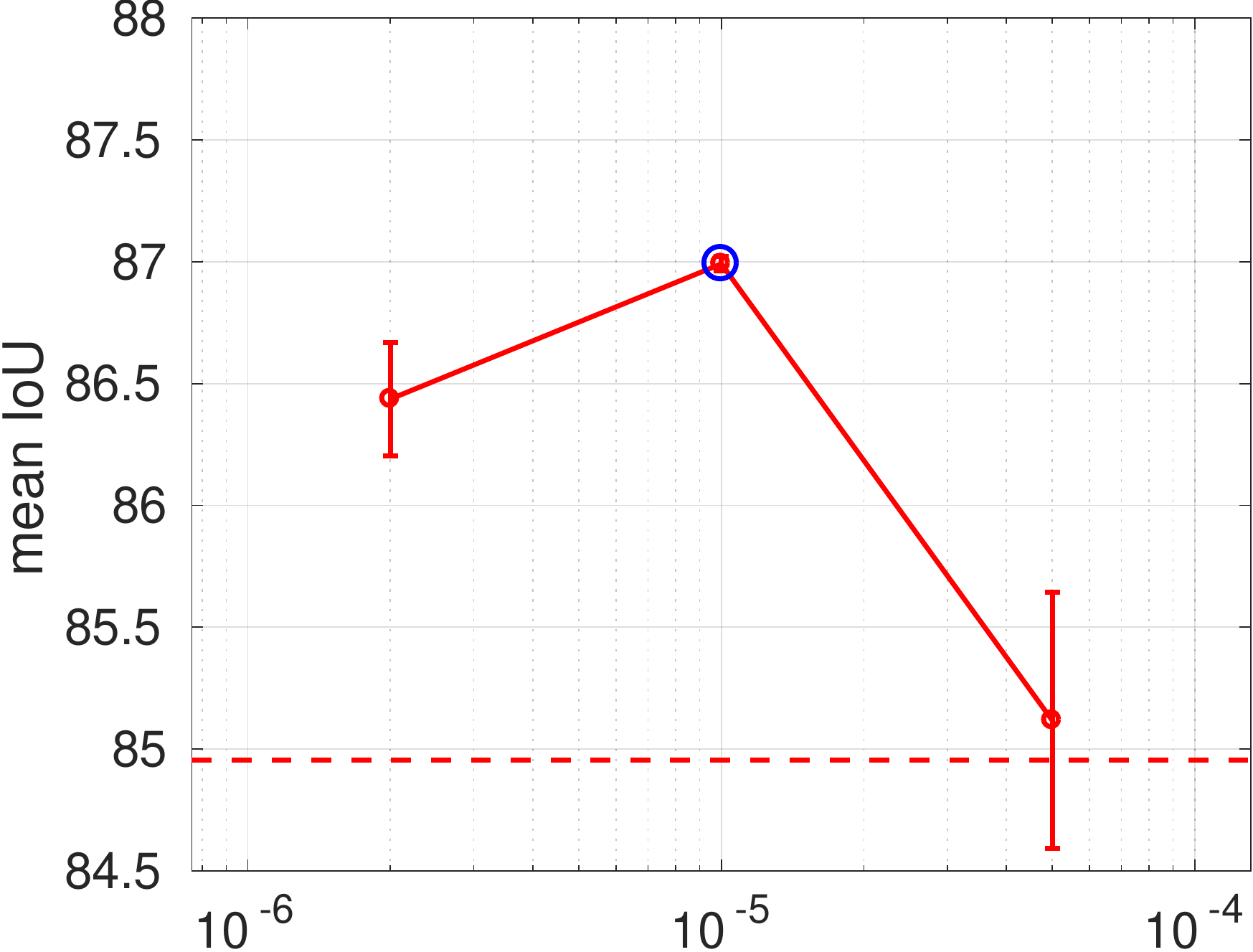}}
\hfill
\subfigure[online loss scale $\beta$]{\includegraphics[scale=0.23]{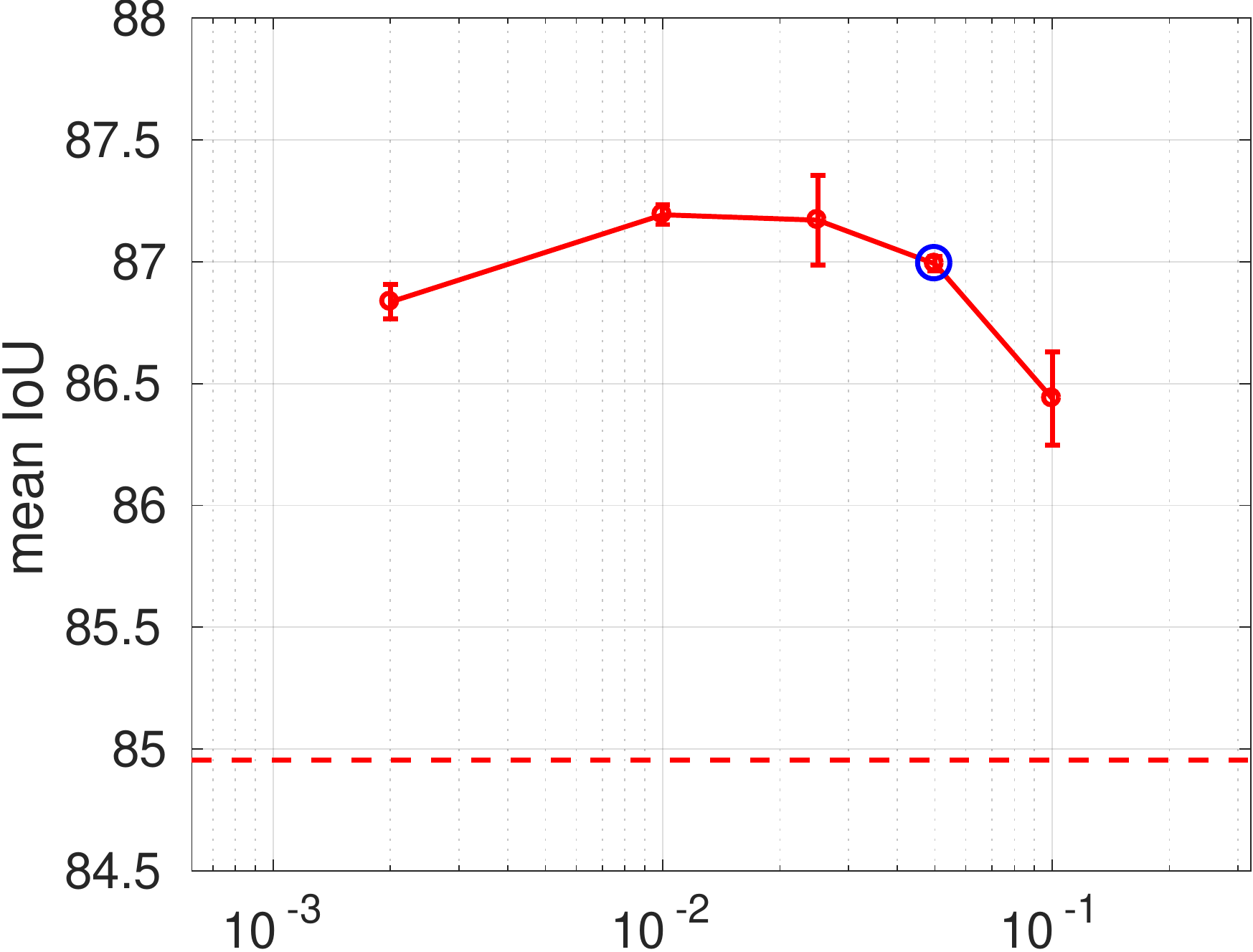}}
\hfill
\subfigure[distance threshold $d$]{\includegraphics[scale=0.23]{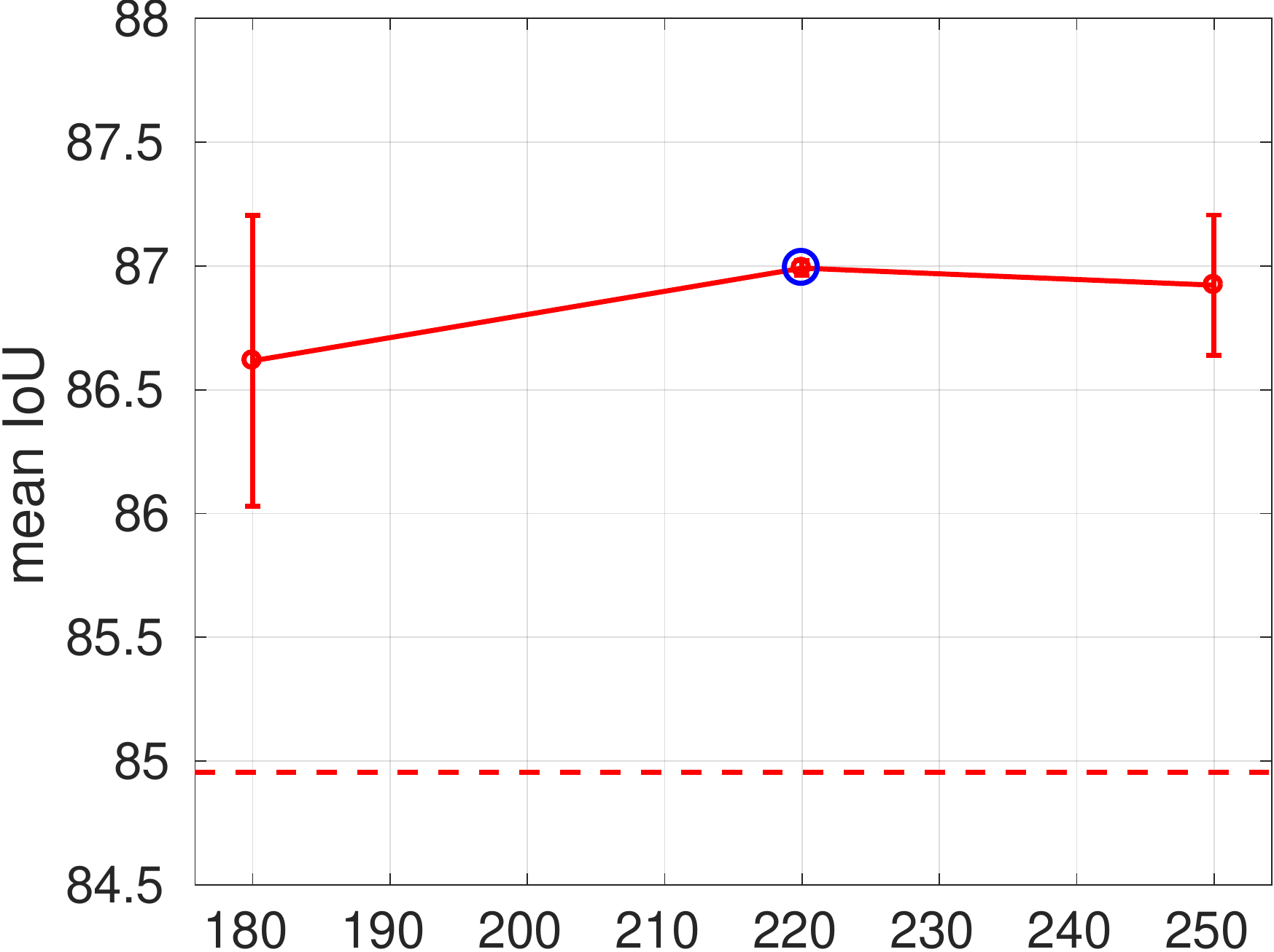}}
\subfigure[total steps $n_{online}$]{\includegraphics[scale=0.23]{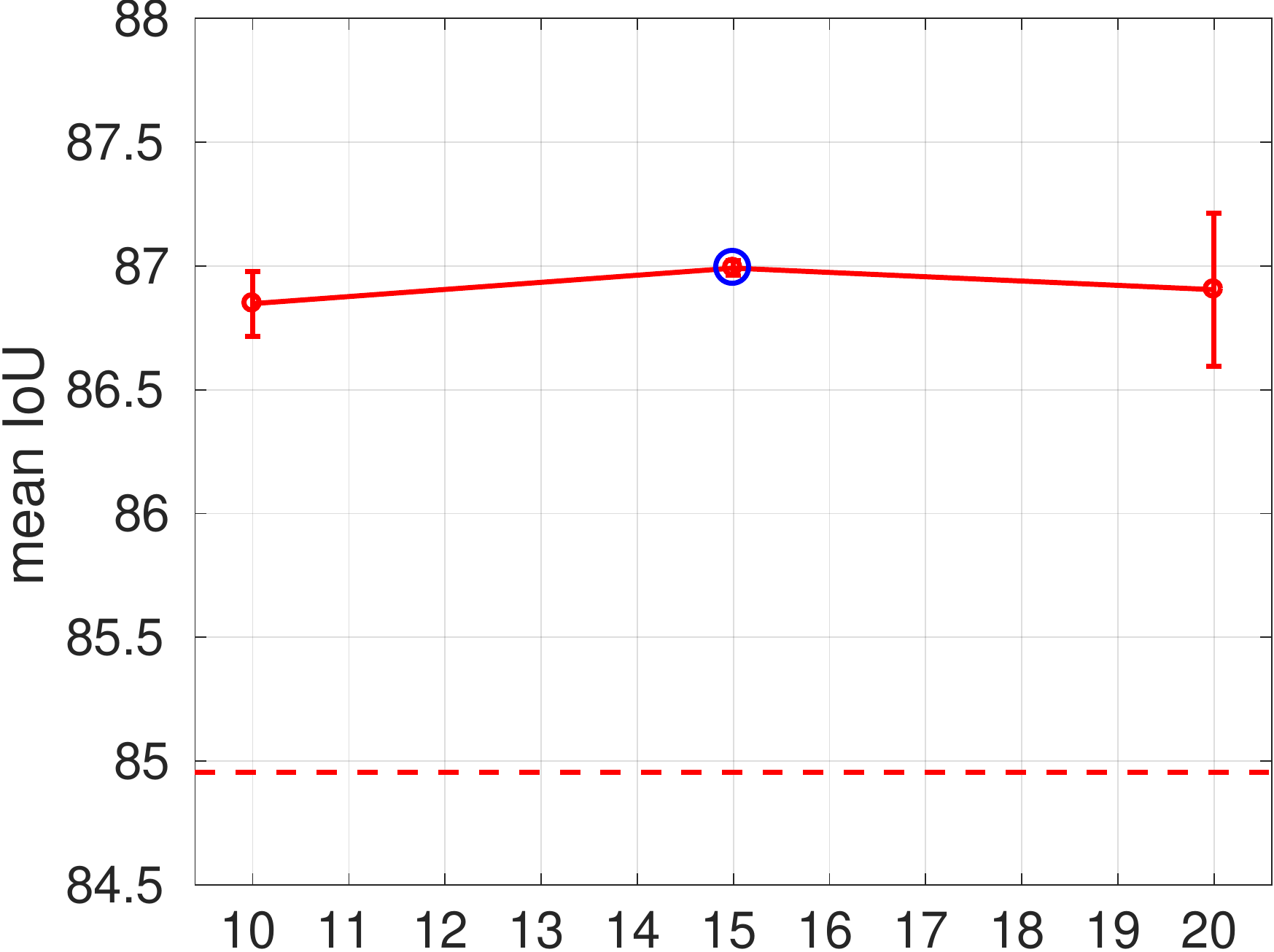}}
\hfill
\subfigure[update steps $n_{curr}$]{\includegraphics[scale=0.23]{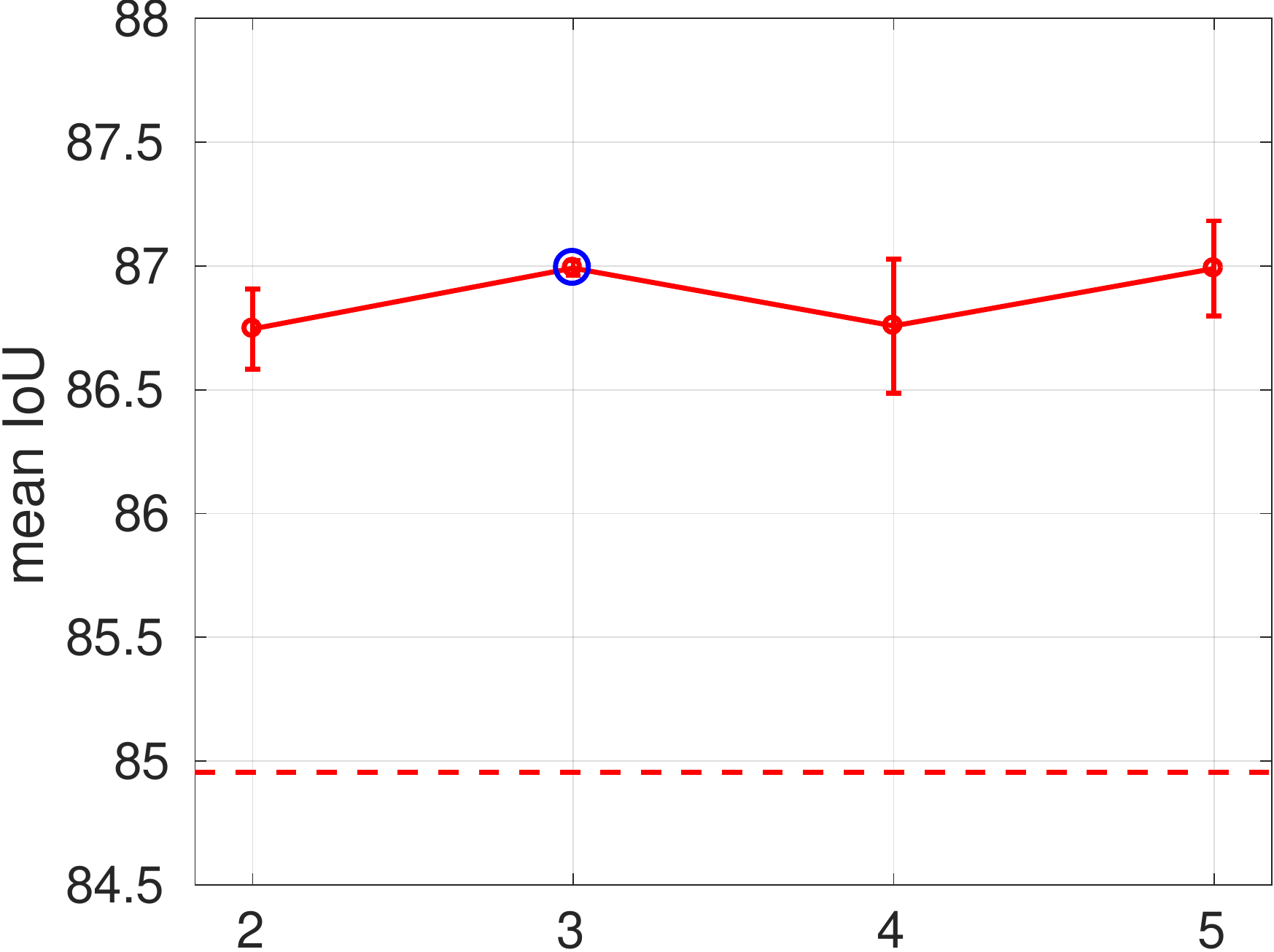}}
\hfill
\subfigure[positive threshold $\alpha$]{\includegraphics[scale=0.23]{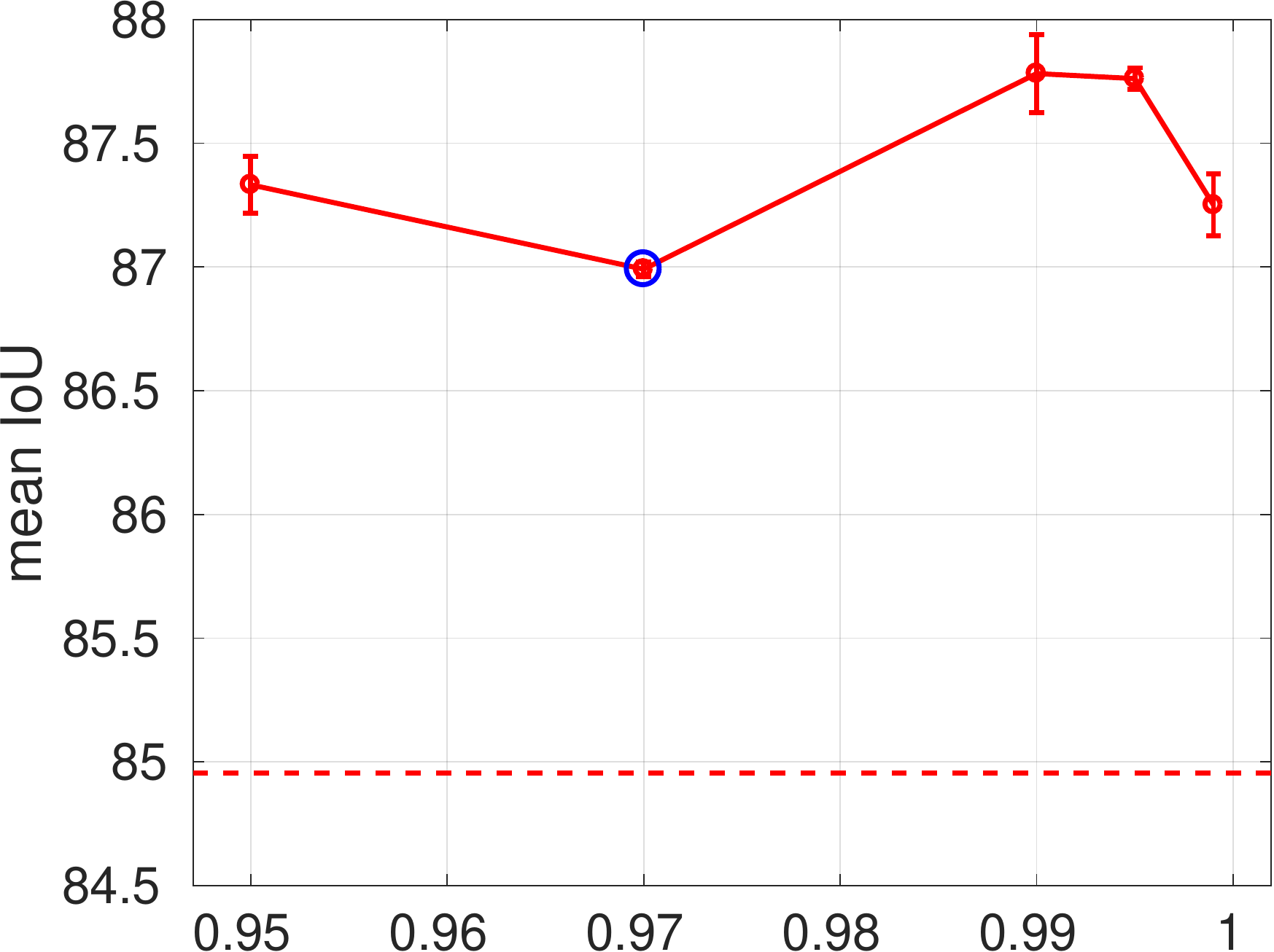}}
\subfigure[erosion size]{\includegraphics[scale=0.23]{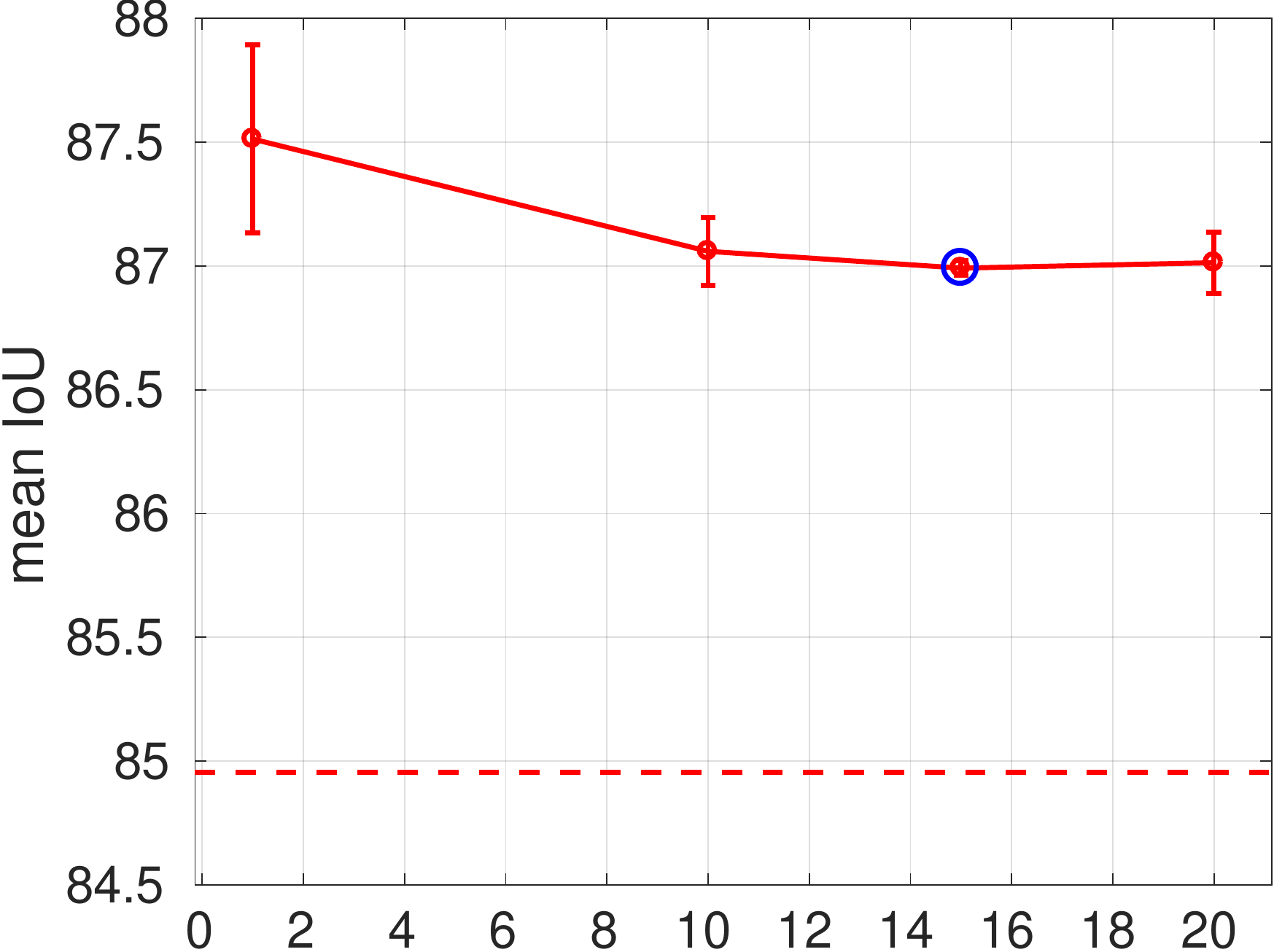}}

\caption{\label{fig:hyperparams}Influence of online adaptation hyperparameters on the DAVIS training set. The blue circle marks the operating point, based on which one parameter is changed at a time. The dashed line marks the un-adapted baseline. The plots show that overall our method is very robust against the exact choice of hyperparameters, except for the online learning rate $\lambda$. The standard deviations estimated by three runs are shown as error bars. In some cases, including the operating point, the estimated standard deviation is so small that it is hardly visible.}
\end{figure}

As described in the main paper, we found $\alpha=0.97$, $\beta=0.05$, $d=220$, $n_{online}=15$, $n_{curr}=3$, $\lambda=10^{-5}$ and $15$ for the erosion size to work well on DAVIS. Starting from these values as the operating point, we conducted a more detailed hyperparameter study by changing one hyperparameter at a time, while keeping all others constant (see Fig. \ref{fig:hyperparams}). The plots show that the performance of \methodname{OnAVOS} is in general very stable with respect to the choice of most of its hyperparameters and for every configuration we tried, the result was better than the un-adapted baseline (the dashed line in the plots). The single most important hyperparameter is the online learning rate $\lambda$, which is common for deep learning approaches. The online loss scale $\beta$ and the positive threshold $\alpha$ have a moderate influence on performance, while changing the distance threshold $d$ and the number of steps $n_{online}$ and $n_{curr}$ in a reasonable range only leads to minor changes in accuracy. For the erosion size, the optimum is achieved at 1, \ie when no erosion is applied. This result suggests that the erosion operation is not helpful for DAVIS. The plots show that there is still some potential for improving the results by further tuning the hyperparameters. However, this study was meant as a characterization of our method rather than a systematic tuning. 

The generalizability and the robustness of \methodname{OnAVOS} with respect to the choice of hyperparameters is further confirmed by the experiments on YouTube-Objects, which used the same hyperparameter settings as on DAVIS.


\end{document}